# Automatic Detection of Texture Defects using Texture-Periodicity and Gabor Wavelets

V Asha[1], N U Bhajantri[2], and P Nagabhushan[3]

[1]New Horizon College of Engineering, Bangalore, Karnataka, India
v_asha@live.com

[2]Govt. College of Engineering, Chamarajanagar, Mysore District, Karnataka, India
bhajan3nu@gmail.com

[3]University of Mysore, Mysore, Karnataka, India
pnagabushan@hotmail.com

**Abstract**

In this paper, we propose a machine vision algorithm for automatically detecting defects in textures belonging to 16 out of 17 wallpaper groups using texture-periodicity and a family of Gabor wavelets. Input defective images are subjected to Gabor wavelet transformation in multi-scales and multi-orientations and a resultant image is obtained in L2 norm. The resultant image is split into several periodic blocks and energy of each block is used as a feature space to automatically identify defective and defect-free blocks using Ward's hierarchical clustering. Experiments on defective fabric images of three major wallpaper groups, namely, pmm, p2 and p4m, show that the proposed method is robust in finding fabric defects without human intervention and can be used for automatic defect detection in fabric industries.

**Key words**: Periodicity, Gabor wavelet, Defect, Cluster.

## 1 Introduction

Product inspection is a major concern in quality control of various industrial products. Textile industry is one of the biggest traditional industries where automated inspection system will help in reduced inspection time and increased production rate. Though patterned fabric designs being produced by modern textile industries are plenty, all patterned fabrics can be classified into only 17 wallpaper groups (denoted as p1, p2, p3, p3m1, p31m, p4, p4m, p4g, pm, pg, pmg, pgg, p6, p6m, cm, cmm and pmm) that are composed of lattices of parallelogram, rectangular, rhombic, square or hexagonal shape [1]. Strictly speaking, p1 defines a texture with just one lattice repeating itself over the complete image such as plain and twill fabrics. Among the other 16 wallpaper groups, pmm, p2 and p4m are called major wallpaper groups as other groups can be transformed into these 3 major groups through geometric transformation [2]. Inspection on patterned textures belonging to wallpaper groups other than p1 group is more complicated than that in textures belonging to p1 group due to complexity in the design, existence of numerous categories of patterns, and similarity between the defect and background [3]. So, most of the methods in literature rely on training stage with numerous defect-free samples for obtaining decision-boundaries or thresholds [3-7]. Moreover, conventional human vision based inspection has the following shortfalls:

- Lack of reproducibility of inspection results due to fatigue and subjective nature of human inspections
- Prolonged inspection time
- Lack of perfect defect detection due to complicated design in fabric patterns manufactured by modern textile industries

Motivated by the fact that the response of the Gabor wavelets is similar to human visual system [8], we make use of Gabor-space of the defective image to discriminate between defect-free and defective zones and propose a method of defect detection on patterned textures belonging to 16 out of 17 wallpaper groups without any training stage. The main contributions of this research are summarized as follows:

- The proposed method is more generic as it can be applied to periodic images belonging to 16 out of 17 wallpaper groups (other than p1 group).
- The proposed method does not need any training stage with defect-free samples for decision boundaries or thresholds unlike other methods.
- Detection of defective periodic blocks is automatically carried out based on cluster analysis without human intervention.

The organization of this paper is as follows: Section-2 gives a brief review on Gabor wavelets, proposed algorithm for defect detection and experiments on various real fabric images with defects. Section-3 has the conclusions.

## 2 Proposed model for defect detection

### 2.1 Gabor wavelets

In visual perception of real-world, Gabor wavelets capture the properties of spatial localization, orientation selectivity, spatial frequency selectivity, and quadrature phase relationship and are good approximation to the filter response profiles encountered experimentally in cortical neurons [8]. Gabor wavelets are widely used for image analysis because of their biological relevance and computational properties and are defined as follows [9]:

$$\psi_{\theta,v}(z) = \frac{k_{\theta,v}^2}{\sigma^2} \exp\left(-\frac{k_{\theta,v}^2 z^2}{2\sigma^2}\right) \left[\exp(ik_{\theta,v}) - \exp\left(-\frac{\sigma^2}{2}\right)\right]. \tag{1}$$

where $\sigma = 2\pi$. The symbols $\theta$ and $v$ represent the orientation and the scale of the Gabor wavelet, respectively in the spatial domain $z = (x, y)$ and the wave vector $k_{\theta,v}$ is given by

$$k_{\theta,v} = k_v \exp(-i\theta). \tag{2}$$

where $k_v = k_{max}/f_v$; $k_{max} = \pi/2$, $f = \sqrt{2}$. The Gabor wavelets are self-similar as these can be generated from one wavelet through scaling and rotation via the wave vector. Each Gabor wavelet is a product of a Gaussian envelope and a complex wave with real and imaginary parts.

Fig. 1 shows typical Gabor wavelets (real and imaginary parts) of size 36 × 27 pixels generated using 5 scales $v \in \{0, 1, 2, 3, 4\}$ and 8 orientations $\theta \in \{0, \pi/8, \pi/4, 3\pi/8, \pi/2, 5\pi/8, 3\pi/4, 7\pi/8\}$.

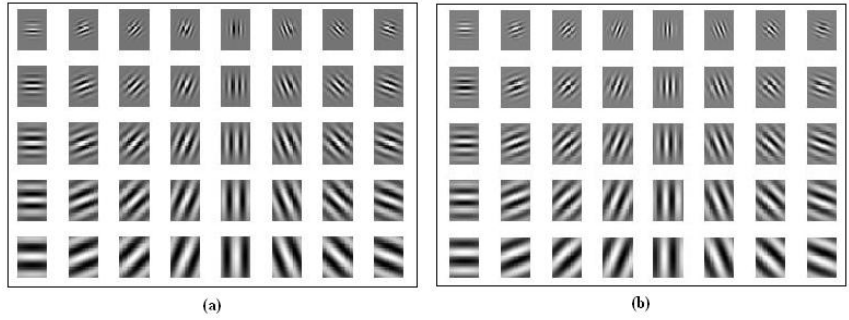

**Fig. 1:** (a) Real Gabor wavelets; (b) Imaginary Gabor wavelets. From left to right, the orientations are 0, π/8, π/4, 3π/8, π/2, 5π/8, 3π/4, and 7π/8 radians and from top to bottom, the scales are 0, 1, 2, 3, and 4 for both real and imaginary wavelets.

### 2.2 Description of the algorithm

There are three main assumptions in the proposed algorithm as follows:
- The image under inspection has at least two periodic units in horizontal direction and two in vertical direction of known dimensions.
- Number of defective periodic units is always less than the number of defect-free periodic units.
- Images under inspection are from imaging system oriented perpendicular to the surface of the product such as textile fabric.

As far as images of p1 wallpaper group (such as plain and twill fabrics) are concerned, there is flexibility in selecting the size of the Gabor kernels [10]. However, in general, more reliable measurement of texture features calls for larger window sizes, whereas, extracting finer details calls for smaller windows [11]. For a periodic patterned texture, size of the filter can be chosen to be same as the size of the periodic unit. For images belonging to wallpaper groups other than p1, there are sub-patterns within a periodic pattern. Hence, for all the test images, size of the kernel is selected to be half the size of the periodic unit and each input defective image is subjected to Gabor wavelet transformation using Gabor kernels in 5 scales $v \in \{0, 1, 2, 3, 4\}$ and 8 orientations $\theta \in \{0, \pi/8, \pi/4, 3\pi/8, \pi/2, 5\pi/8, 3\pi/4, 7\pi/8\}$ to get a resultant image in L2 norm [12]. Since the proposed method is periodicity-based, from the resultant Gabor filtered image of size $M \times N$, four cropped images of size $M_{crop} \times N_{crop}$ are obtained by cropping the resultant image from all 4 corners (top-left, bottom-left, top-right and bottom-right). Size of each cropped image ($M_{crop} \times N_{crop}$) is calculated using the following equations:

$$M_{crop} = floor\ (M \times P_c) \times P_c \qquad (3)$$

$$N_{crop} = floor\ (N \times P_r) \times P_r \qquad (4)$$

where $P_r$ is the row periodicity (i.e., number of columns in a periodic unit) and $P_c$ is the column periodicity (i.e., number of rows in a periodic unit). Each cropped image is split into several periodic blocks and energy of each block in L1 norm [12] is used as a feature space for Ward's hierarchical clustering [13] to get defective and defect-free clusters.

**2.3 Illustration of the algorithm, experiments and results**

In order to illustrate the proposed method of defect detection, let us consider a defective image (p4m image with thick bar defect) as shown in Fig. 2 (a). The resultant image after Gabor wavelet transformation using Gabor kernels (in 5 scales and 8 different orientations) is shown in Fig. 2 (b). Defective and defect-free blocks are automatically identified from all cropped images obtained from the Gabor filtered image using Ward's clustering with energy of periodic blocks as feature space as shown in Fig. 2 (c)-(f).

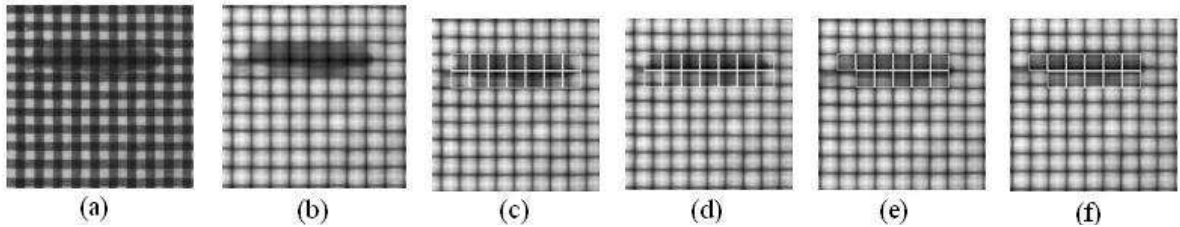

**Fig. 2:** (a) p4m test image with thick bar defect; (b) Gabor-space of the image; (c), (d), (e) and (f) highlight the boundaries of the defective periodic blocks (in white pixels) identified from the cropped images.

In order to get the overview of the total defects in the input image, we use fusion of defects. The boundaries of the defective periodic blocks identified from each cropped image when superimposed on the Gabor-space of the original image (as shown in Fig. 3 (a)) can give an overview of the total defects. However, in order to extract the edges of the total defects, boundaries of the defective periodic blocks are taken on a plain background as shown in Fig. 3 (b) and are morphologically filled [14] to get fused defective zones as shown in Fig. 3 (c). Edges of these fused defects are extracted using Canny edge operator [14] and are superimposed on the original defective image to get the overview of the total defects in the input image as shown in Fig. 3 (d). Real fabric images of 3 major wallpaper groups (pmm, p2 and p4m) with defects such as broken end, hole, thin bar and thick bar are also tested using the proposed method. Fig. 4 shows the final test results from experiments on these defective fabric images. In order to access the performance of the proposed method, performance parameters [15], namely, precision, recall and accuracy are all calculated in terms of true positive (TP), true negative (TN), false positive (FP), and false negative (FN), where true positive is the number of defective periodic blocks identified as defective, true negative is the number of defect-free periodic blocks identified as defect-free, false positive is the number of defect-free periodic blocks identified as defective and false negative is the number of defective periodic blocks identified as defect-free. Precision is calculated as TP/(TP+FP). Recall is calculated as TP/(TP+FN). Accuracy is calculated as (TP+TN)/(TP+TN+FP+FN). The average metrics (precision, recall and accuracy), calculated for pmm, p2 and p4m groups, are (100%, 68.6% and 93.9%), (100%, 90.2% and 99.6%), and (100%, 78.9% and 99.0%) respectively (based on total number of samples - 864, 2640 and 1440 for pmm, p2 and p4m groups). Relatively less recall rates indicate that there are few false negatives identified by the proposed method. However, since the proposed method yields high precision and accuracy, the proposed method can contribute to automatic defect detection in fabric industries.

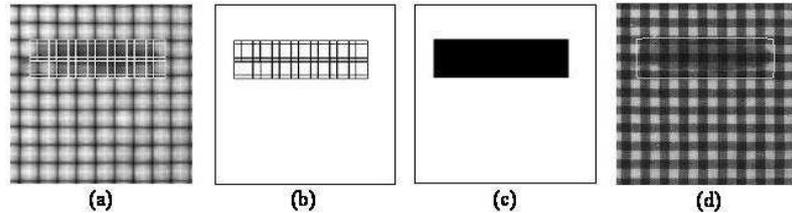

**Fig. 3.** Illustration of defect fusion: (a) Boundaries extracted from all cropped images shown superimposed on the Gabor-space of the original defective image; (b) Boundaries extracted from all cropped images shown separately on a plain background; (c) Result of morphological filling; (d) Edges of the fused defective blocks identified using Canny edge operator and shown superimposed on the original defective image.

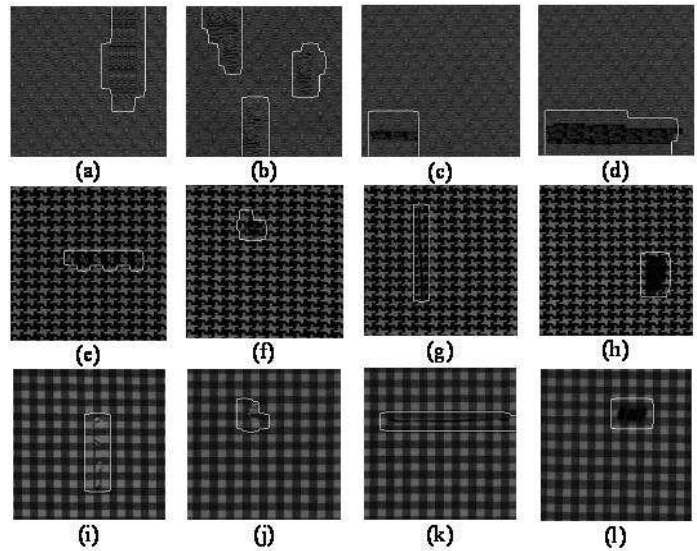

**Fig. 4:** Result of the proposed method of defect detection on real fabric images. First, second and third rows show pmm, p2 and p4 images respectively. Defects in first, second, third and fourth columns are broken end, hole, thin bar and thick bar respectively.

## 3 Conclusions

In the proposed method of defect detection, texture-periodicity has been effectively utilized for determining the size of the Gabor kernels for each test image and for analyzing the defective blocks in Gabor-space using hierarchical clustering. Fusion of defects identified from four cropped images generated from the input image helps in getting an overview of the total defects. Experiments on defective fabric images that belong to three major wallpaper groups (pmm, p2 and p4m) illustrate that the proposed method can contribute to the development of computerized defect detection in fabric industries.

*Acknowledgment*

The authors would like to thank Dr. Henry Y. T. Ngan, Research Associate of Industrial Automation Research Laboratory, Department of Electrical and Electronic Engineering, The University of Hong Kong, for providing the database of patterned fabrics.